\documentclass[runningheads]{llncs}
\usepackage{graphicx}
\usepackage{tikz}
\usepackage{comment}
\usepackage{amsmath,amssymb} % define this before the line numbering.
\usepackage{mathtools}
\usepackage{dsfont}
\usepackage{color}

\usepackage{algorithm}% http://ctan.org/pkg/algorithm
\usepackage{algpseudocode}% http://ctan.org/pkg/algorithmicx
\usepackage{subcaption}
\usepackage[accsupp]{axessibility} % Improves PDF readability for those with disabilities.

\usepackage{fancyhdr}% http://ctan.org/pkg/fancyhdr
\begin{document}

% \linenumbers
\pagestyle{headings}
\mainmatter
\def\ECCVSubNumber{16}  % Insert your submission number here

\title{Matching Multiple Perspectives for Efficient Representation Learning}

%\begin{comment}
% \titlerunning{ECCV-22 submission ID \ECCVSubNumber} 
% \authorrunning{ECCV-22 submission ID \ECCVSubNumber} 
% \author{Anonymous ECCV submission}
% \institute{Paper ID \ECCVSubNumber}
%\end{comment}
%******************

%\begin{comment}
% \titlerunning{Matching Multi-Perspective Robot Views for Self-Supervised Learning}
% \titlerunning{Efficient Representation Learning by Matching Multi-Perspective Views}
% \titlerunning{Matching Multi-Perspective Views for Efficient Representation Learning}
\titlerunning{Matching Multiple Perspectives for Efficient Representation Learning}
\author{Omiros Pantazis\inst{1*} \hspace{0.2cm}
Mathew Salvaris\inst{2}}
\authorrunning{Pantazis et al.}
\institute{University College London \and iRobot}
% % \email{lncs@springer.com}\\
%\end{comment}
%******************
\maketitle

\begin{abstract}
Representation learning approaches typically rely on images of objects captured from a single perspective that are transformed using affine transformations. Additionally, self-supervised learning, a successful paradigm of representation learning, relies on instance discrimination and self-augmentations which cannot always bridge the gap between observations of the same object viewed from a different perspective.  Viewing an object from multiple perspectives aids holistic understanding of an object which is particularly important in situations where data annotations are limited. In this paper, we present an approach that combines self-supervised learning with a multi-perspective matching technique and demonstrate its effectiveness on learning higher quality representations on data captured by a robotic vacuum with an embedded camera.
We show that the availability of multiple views of the same object combined with a variety of self-supervised pretraining algorithms can lead to improved object classification performance without extra labels.
% \keywords{We would like to encourage you to list your keywords within
% the abstract section}
\end{abstract}

\thispagestyle{fancy}%
\section{Introduction}

\begin{figure}[t]
\begin{center}
 \includegraphics[width=0.6\columnwidth]{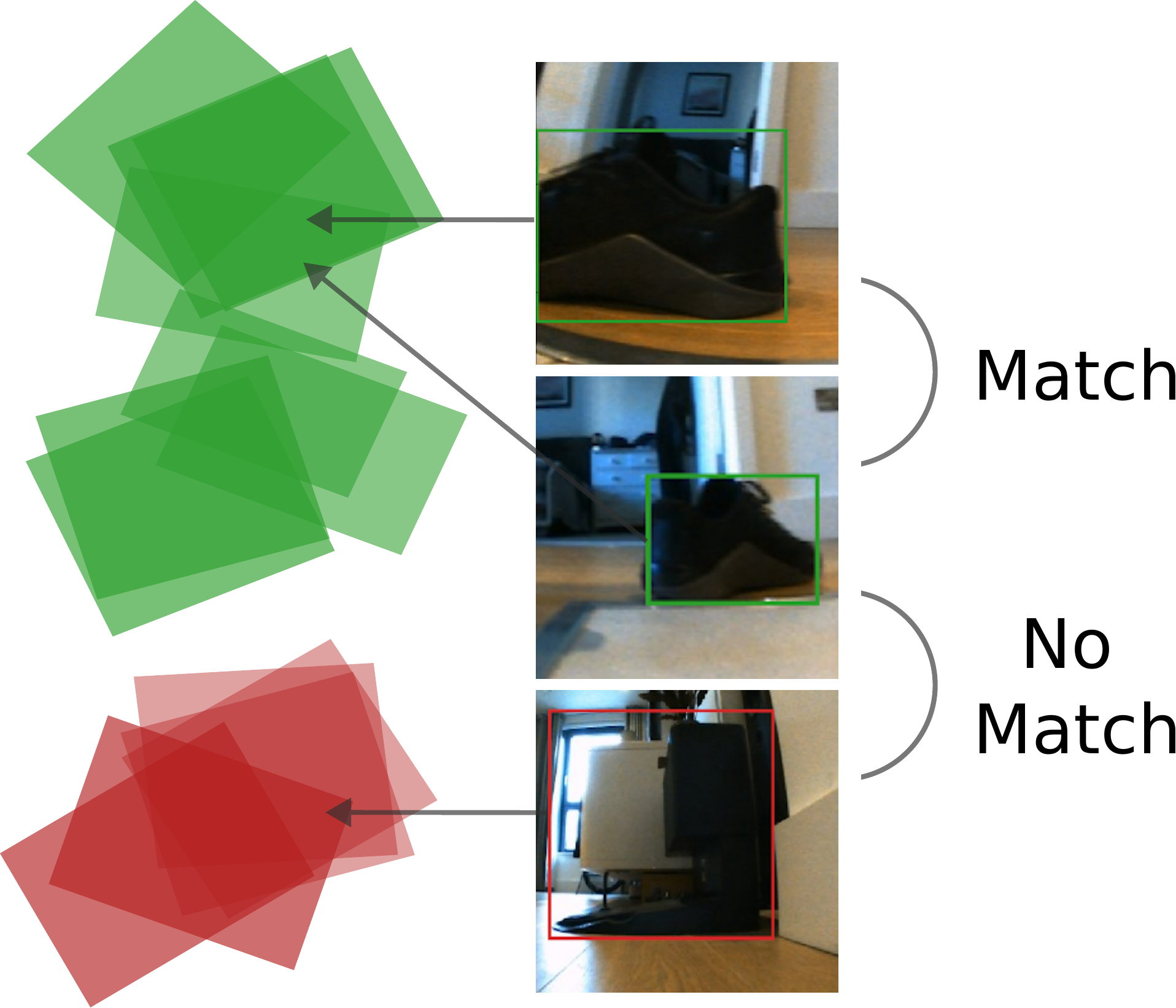}
\end{center}
  % \vspace{-10pt}
   \caption{The agent with the embedded camera traverses the space and captures images. The images with overlapping projected polygons in the room map represent different views of the same object (match) and can be used as pairs for self-supervised learning. Here we illustrate views of shoes (green) and a robot dock (red) observed from various perspectives.
   }
  % \vspace{-15pt}
\label{fig:polygon_match}
\end{figure}

Mobile robots are increasingly asked to recognize objects to inform their decision making process under a variety of real-world situations. For example, a home robot may need to automatically identify obstacles to stay clear of them and more critically a self-driving car must identify road signs without fault\cite{swaminathan2019autonomous}, whereas the above challenges can present themselves under different lighting conditions, levels of view obstruction or difficult angles. Having said that, robots nowadays have started to rely on embedded cameras and deep learning systems that help them navigate \cite{zhu2021deep} and discriminate between targets or obstacles \cite{skoczen2021obstacle}. Of course, there are challenges arising from the computational restrictions of the hardware, the fact that the agent is moving while taking pictures and the endless variation in conditions presented across natural environments. Tackling these difficulties while solving challenging visual tasks would typically demand a long and expensive annotation procedure and despite the fact that the emergence of benchmark datasets \cite{deng2009imagenet,lin2014microsoft} enabled rapid progress in computer vision, neither generalization from them nor repeating their annotation procedure in every real-world task dataset are options. Thus, learning  transferable representations of visual data without requiring explicit semantic supervision at training time is an important and open problem in computer vision.  

On that note, efficient representation learning has been a key object of studies in the recent years, with multiple self-supervised learning (SSL) methods reporting remarkable performance while using a fraction of the data labels \cite{He_2020_CVPR,chen2020simple,grill2020bootstrap,caron2020unsupervised,dwibedi2021little,zbontar2021barlow}. The first wave of successful SSL techniques that were able to compete with fully-supervised benchmarks involved a pretext task that produces high quality representations by forcing invariance to a predefined set of transformations \cite{He_2020_CVPR,chen2020simple,grill2020bootstrap}. This is mainly achieved by forming pairs of images and their augmented versions (i.e. creating positive pairs) and pushing them together through a loss function. However, subsequent works devised ways to retrieve positives that turned to be more informative for self-supervised pretraining when compared with traditional self-augmentations \cite{dwibedi2021little,ayush2020geography,Pantazis_2021_ICCV,azizi2021big}. For example, in \cite{dwibedi2021little}, the authors mined positives based on the image’s nearest neighbours in representation space. Moving away from ImageNet, beneficial and diverse positives were also mined by exploiting the contextual information that ordinarily comes with naturally collected data. More specifically, \cite{ayush2020geography,Pantazis_2021_ICCV} take advantage of the spatiotemporal information that comes with images collected from satellites or wild cameras to mine effective positives for the self-supervised pretext task. In another work \cite{azizi2021big}, positive pairs were formed through the availability of multiple photos for a patient’s medical case. Natural variation in the self-supervised learning pair formation can also be introduced in sequential frames by pushing together frames that have temporal proximity within a video \cite{sermanet2018time,purushwalkam2020demystifying,qian2021spatiotemporal}. The common denominator of the aforementioned successful positive retrieval techniques is that a static viewpoint (camera) is able to capture a dynamic set of views and a subset of these views can be linked in some cases through the availability of metadata. Nevertheless, there are numerous scenarios where a dynamic agent collects data about a static environment. These scenarios can be encountered in cameras that are embodied in agents (robots) and can range from robotic vacuums and agricultural robots to space and ocean exploration robots \cite{zereik2018challenges,arm2019spacebok,skoczen2021obstacle}. To the best of our knowledge, there is no exploration of self-supervised representation learning for object classification where the viewpoint is dynamic and the objects of interest are static, and we attempt to demonstrate and tackle such a scenario with the use-case of this paper.

The fact that moving agents with embedded cameras capture data while navigating through an environment enables collection of all-around views of objects. However, simply using meta-information such as 2-Dimensional coordinates or time cannot ensure retrieval of positives that come from the same class. For this reason, we exploit the camera intrinsics and extrinsics as well as the robot position to gain an understanding of  a) where the agent is located within the environment, b) where the agent is looking and finally for c) mapping the 2-D images to the 3-Dimensional space. The aforementioned mapping and camera localization enables us to understand whether two boxes that come from a different photo, enclose the same object. This exercise will provide for each image or detected box within an image, a list of candidate images that can be selected to form a positive pair in the self-supervised pretext task. An illustration of this process can be observed in Figure \ref{fig:polygon_match}. The assumption is that by pushing together embeddings that come up from different perspectives of the same object we can form a SSL task that leads to better representations. We test our assumption in data collected from a robotic vacuum across multiple episodes, where each episode corresponds to a set of photos captured within the same house through a period of time. The downstream task we employ for evaluation is object classification, where each class corresponds to an object the vacuum needs to avoid.

The key contribution of this paper is twofold: 
\begin{enumerate}

    \item We show how recent self-supervised learning techniques can assist image classification in a challenging dataset with images collected in a home environment by a navigating robot. Exploitation of self-supervised learning proved to be more effective than transfer learning from an ImageNet pretrained network.
    
    \item We propose Polygon Matching, a novel way to mine informative positives for self-supervised learning in data collected from robot agents. We utilize our approach on top of three different self-supervised learning techniques and report consistent gains in accuracy.
   
\end{enumerate}

\section{Related Work}

\subsection{Self-Supervised Learning}

Self-supervised learning for computer vision applications can be broken down into pretext learning, where a model is trained on unlabeled image collections given a self-supervisory signal and the downstream task where the representations learnt from the pretext task are exploited in tandem with available annotations.
Earlier works in self-supervised representation learning involved the meticulous design of pretext tasks that are capable to generate supervisory signal out of unlabeled image collections \cite{doersch2015unsupervised,noroozi2016unsupervised,zhang2016colorful,gidaris2018unsupervised}. Then, researchers were able to successfully exploit strong image augmentations to generate alternative views of images that are capable of introducing visual variance while preserving the key content of images \cite{He_2020_CVPR,chen2020simple,grill2020bootstrap,zbontar2021barlow}. At first, suggested approaches used contrastive learning \cite{He_2020_CVPR,chen2020simple,dwibedi2021little} to push together the embeddings that are associated with a different augmentation while pushing them away from the embeddings of other images in a memory bank \cite{He_2020_CVPR} or within the same batch \cite{chen2020simple,dwibedi2021little,zbontar2021barlow}. The success of the contrastive learning framework in SSL led researchers to also exploit it and record notable gains in the supervised domain \cite{khosla2020supervised}. Subsequent works in SSL, produced networks that do not rely on negative pairs \cite{grill2020bootstrap,chen2021exploring} and proved capable of learning high quality representations that lead to high downstream task performance. In our experiments, we use the simple contrastive learning framework (SimCLR) proposed by \cite{chen2020simple} as the basis of our multi-perspective mining approach because it is intuitive and quite representative of most of the recent state-of-the-art techniques. In addition, we also explore SimSiam \cite{chen2021exploring} as a representative of approaches that do not use negatives and a simple Triplet loss \cite{weinberger2009distance}.

Most of the aforementioned self-supervised approaches benchmark their contributions on ImageNet \cite{deng2009imagenet}, but the successes of self-supervised learning are not limited there. Self-supervised learning also proved to be effective on learning high quality representations of data across a variety of challenging tasks such as species classification from camera traps \cite{Pantazis_2021_ICCV}, medical image analysis \cite{azizi2021big} or remote sensing from satellite data \cite{ayush2020geography}. Similarly in our work, we use a challenging dataset comprising of images collected from the embedded camera of a robotic agent while traversing home environments.

\subsection{Positive Mining}

The relationship between self-supervised learning and various real-world applications has not been restricted to simple application of the most advanced self-supervised algorithms. On the contrary, the contextual information that comes from real-world imagery proved to be suitable for the properties of self-supervised learning, leading to further improvements. Indicatively, multiple medical images taken from a patient on the same day \cite{azizi2021big}, spatio-temporal proximity of camera trap frames \cite{Pantazis_2021_ICCV}, or exploitation of consecutive frames within videos \cite{sermanet2018time,purushwalkam2020demystifying,qian2021spatiotemporal,wu2022self} proved to be informative proxies to mine positives for the pretext learning stage. In the aforementioned cases, positive mining led to higher performance in the respective downstream tasks by using widely available metadata to introduce variation during pretext learning, which could not simply be introduced by self-augmentations. Even though, these scenarios help deepen the variability within the contrastive learning task, they lack variation in perspective that stems from a different point-of-view or lighting conditions that may be associated with different parts of the day.

Here, we propose an approach that exploits the positioning of an agent with an embedded camera within a room along with the direction of the camera to figure out what subspace of the room is captured by the photo. This information is exploited to construct positive pairs of images for self-supervised learning, where positives are defined as images that capture similar points in space. With the above, we aspire to learn better representations by bringing closer embeddings of different perspectives of each object.

\section{Multi-Perspective Views in Self-Supervised Learning}

\subsection{Simple Contrastive Learning framework and Variants}

Our approach is built on top of SSL approaches that exploit instance discrimination as signal for efficient representation learning. Indicatively, SimCLR \cite{chen2020simple}, one of the approaches we explore, is a simple contrastive self-supervised framework that for each image $x_i \in \mathcal{X}$ uses its strongly augmented version ${x}_p$ as its positive pair and the rest of the images ${x}_n$ in the same batch as negatives. Specifically, an image $x_i$ and its transformation go through a feature extractor $f$ (e.g.\ ResNet \cite{he2016deep}) and then through a Multilayer Perceptron (MLP) $g$ that projects them into a lower-dimensional embedding vector $z_i = g(f(x_i))$.

Then, learning takes place through a normalized temperature-based softmax cross-entropy loss \cite{chen2020simple} that is defined for every image ${x}_i$ within a batch $\mathcal{B}$ as
\begin{equation}
 L_{i}= - \log \left( \frac{\exp({sim(z_i,z_p)/ \tau})}{\sum_{n \in \mathcal{B}}\mathds{1}_{[n\neq i]} exp({-sim(z_i,z_n)/ \tau}) } \right),
\end{equation}

where $sim(.)$ is the cosine similarity between the projected image embeddings, temperature $\tau$ regulates the scale of the distances and $\mathds{1}_{[n\neq i]}$ ensures that each image embedding is not compared with itself. Basically, the loss function $L$ pushes the embedding of the query image close to the embedding of its augmented self while repulsing it from the negatives. The above procedure teaches the network to be invariant to appearance variations that do not affect the key content of the image.

We also examine the more recent SimSiam \cite{chen2021exploring} that maintains satisfactory performance and avoids learning a degenerate solution without the need for negatives, by using $h$, an additional predictor MLP between the projections $z_i$ of the different augmentations of each image. In addition, we also try a simple triplet loss \cite{weinberger2009distance} that selects positives in a similar way while randomly sampling a negative for each query image.

\subsection{Polygon Matching}
\label{sec:pmatch}
We build our task on top of the aforementioned SSL approaches, which we also compare against. The difference is on the images that are considered positive pairs and are pushed close to each other during the self-supervised task. We propose a mechanism that aspires to add variation in the signal used as supervision for SSL by selecting positives that cover different perspectives of the same object. 

\noindent{\bf Bounding Box Extraction.} First, we extract bounding boxes from the photo captured by the robot by using a pretrained object detection network, in this case CenterNet \cite {zhou2019objects} that has been trained on a separate dataset. One could easily replace it with a class agnostic network such as from \cite{zhou2021probablistic}. We run inference on the images using this from a run in a number of environments.

\begin{algorithm}
\caption{Polygon Matching Algorithm}\label{pmalgo}
\begin{algorithmic}[1]
\For{\texttt{each Agent}}
    \For{\texttt{each Image $x_i$ captured by the Agent}}
    \State \texttt{Gather robot position, camera parameters ($K, \left[ R \vert t \right]$) and polygon}
    \State \texttt{Project polygon from image plane  $ x_i \in \mathbb{R}^2 $ to world frame  $x_w \in \mathbb{R}^3 $}
    \EndFor
    
    \For{\texttt{each Bounding Box}}
        \State \texttt{Use world map view to find bounding boxes that overlap with it}
        \State \texttt{Record images with overlapping bounding boxes (to be retrieved during SSL as positives)}
    \EndFor
    
\EndFor
\end{algorithmic}
\end{algorithm}

\noindent{\bf Polygon Projection and Matching.} We know the robot's position in the 3D space from the VSLAM system \cite{taketomi2017visual} employed by the robot. We then simply project the bounding box on to the floor based on Inverse Perspective Mapping \cite{mallot1991inverse}. So if viewed from a birds-eye view, the floor would be littered with various polygons that correspond to the individual images, many of which would overlap. To take the polygon from the image plane and project it into the real world we need four pieces of information, the camera intrinsics ($K$) and extrinsics ($ [ R \vert t ]$), robot position and finally the polygon coordinates in the image. The process by which the information is used to project from the robot camera frame to the map frame (or bird's eye view) is briefly described below.

In order to transform the homogeneous image coordinates $ x_i \in \mathbb{R}^3 $ to the homogeneous world coordinates $ x_w \in \mathbb{R}^4 $ we use the projection matrix $ P \in \mathbb{R}^{3 \times 4} $ as given by:
\begin{equation}
  x_i = P x_w \,.
  \label{eq:World2ImageProjection}
\end{equation}
With the projection matrix $P$ encoding the camera's intrinsic parameters $K$ and extrinsics (rotation $ R $ and translation $ t $ with respect to the world frame):
\begin{equation}
  P = K [ R \vert t ] \,.
  \label{eq:ProjectionMatrix}
\end{equation}
Assuming there is a transformation $ M \in \mathbb{R}^{4 \times 3} $ that transforms from the floor plane $ x_f \in \mathbb{R}^3 $ to the world frame
\begin{equation}
  x_w = M x_f \,,
  \label{eq:Road2WorldProjection}
\end{equation}
we can obtain a transformation from image coordinates to the floor plane with the following:
\begin{equation}
  x_f = \left( P M \right)^{-1} x_i \,.
\end{equation}

Then, we check if the polygons that are projected in the map with the aforementioned process have an overlap. The assumption is that the projected bounding boxes that overlap will likely correspond to the same object captured from various viewpoints. Thus, instead of formulating a self-supervised signal by just pushing together an image with an augmented version that arises through a limited combination of transformations, we explore the natural variation that exists in the world by looking at the same object from a different perspective. In practice, for each image $x_i \in \mathcal{X}$ we store the images they overlap with and retrieve them during the SSL task as their positive pair ${x}_p$. The proposed positive retrieval method is summarized in Algorithm \ref{pmalgo}. It is fair to assume that the suggested method, as a positive mining mechanism, can be considered orthogonal to the underlying self-supervised method and thus, robust against future methodological developments in the field. This has been confirmed by preceding positive mining approaches that have consistently reported additive gains under various self-supervised settings and datasets \cite{Pantazis_2021_ICCV}.

\section{Use-Case: Robotic Vacuum Agent}

\begin{figure}[h]
\begin{center}
 \includegraphics[width=0.9\columnwidth]{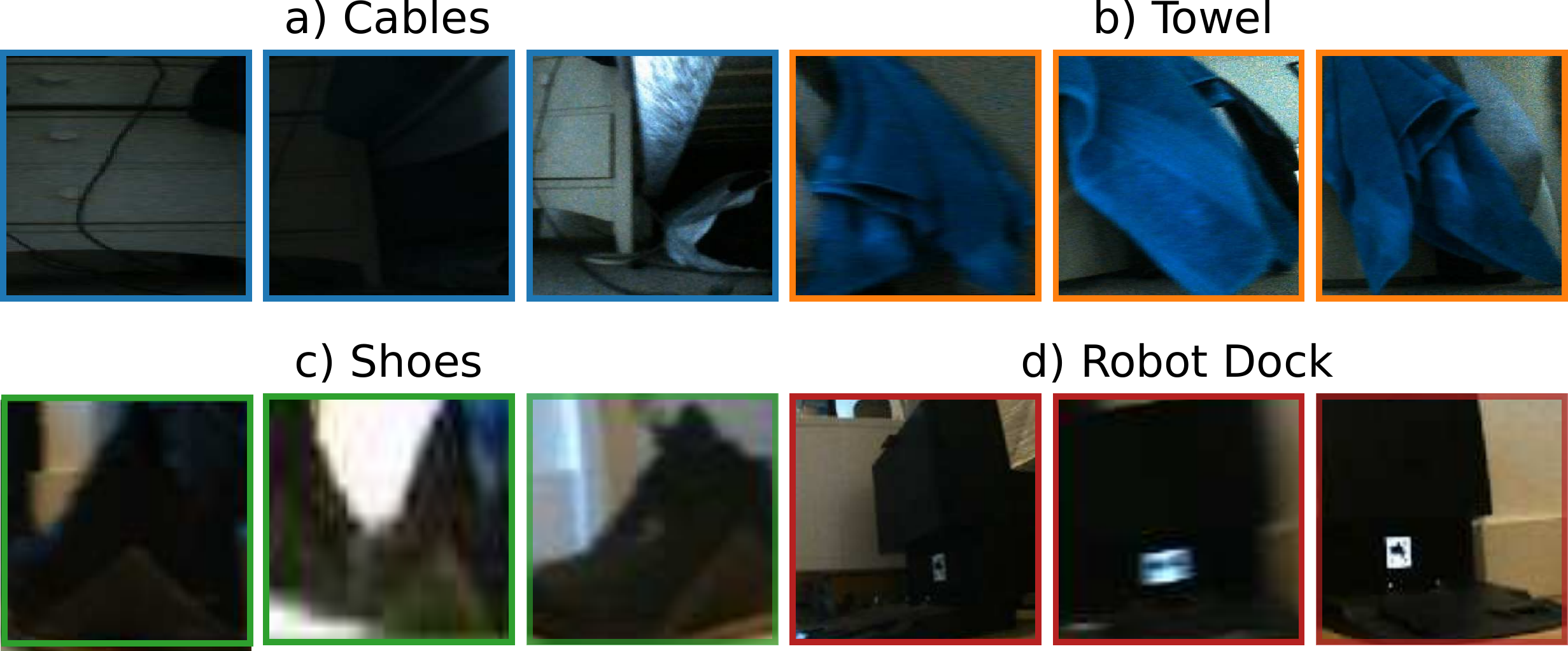}
\end{center}
   \caption{Here we show some images of the \emph{Vaccum Objects} use-case we examined in this work. Specifically, we see multiple images of the same cables, towel, shoes and robot dock as captured from various perspectives. The variety among the different views of the same object stress the need for an objective function that bridges these differences.}
\label{fig:vacuum_images}
\end{figure}

The scenario in which the data is collected for this work is that of a robot vacuum navigating around a house and cleaning. The trajectories of the robot are not modified in any way and therefore the viewpoints of the object are coincidental with the robots path and not planned in any way. During robot operation we use a VSLAM system that localises the robot in the environment. Through this localisation as well as camera projection we can identify whether the images contain the same object or not. The dataset has been collected from hundreds of homes of iRobot employees and therefore includes a considerable amount of variation. Data is collected at around 10Hz at a resolution of 640 by 480. For the purpose of the paper, we refer to this dataset as \emph{Vacuum Objects}. The \emph{Vaccum Objects} dataset covers 12 common objects that can be found inside a typical home environment and the vacuum needs to avoid while it navigates and cleans the house. These include among others: cables, socks, shoes, clothes and pet waste. The challenges of the dataset can be realized by the variety between different views of the same object illustrated in Figure \ref{fig:vacuum_images}. In standard operation the robot may see the same object from 3 to 20 times in a specific mission. It should be noted that the dataset used for these experiments cannot be released but a similar setup can be achieved given any agent with embedded camera that collects data by navigating through space or with simulators such as Meta's Habitat platform \cite{savva2019habitat}. 

\section{Experiments}

\subsection{Implementations Details}

The input images correspond to detections around the object of interest and are acquired using a pretrained CenterNet detector \cite{zhou2019objects} without the use of any specific labels for the \emph{Vacuum Objects} dataset. The CenterNet model was trained using the standard CenterNet hyperparameters \cite{zhou2019objects}. The model was trained using previously gathered data where 12 classes were labeled. In total, there are 261,695 training image boxes generated by CenterNet which we use for self-supervised pretraining and downstream task tuning and 27,626 manually annotated boxes that constitute our test set.

As a pre-processing step, for each image collected with the robot-embedded camera during its navigation, we build a set of matching images using the polygon matching approach described in Section~\ref{sec:pmatch}, i.e. photos collected from the same agent whose polygons intersect with the given image. The image pairings generated in this step are later on retrieved and used during the SSL step, serving as candidate positives for their paired image. This step also involves a depth parameter that defines the maximum depth away from the robot that we consider polygons for, e.g. if an object is too far then its projected polygon will not be taken into account as candidate positive for self-supervised learning. Unless specified differently, the experiments in the paper use 0.7 as the maximum meters away from the camera that we consider.

A ResNet18 \cite{he2016deep} convolutional neural network is used as the backbone feature extractor $f$ for both the self-supervised pretext and the supervised downstream tasks. Across all the SSL approaches tested, the ResNet18 feature extractor is followed by an MLP projector $g$ that maps the features to lower dimensions (128) while for SimSiam we also append an additional predictor MLP $h$ as described in the paper \cite{chen2021exploring}. The input image size we used for the experiments of this paper is $112\times112$. Similar to SimCLR \cite{chen2020simple}, we introduce variance by using transformations such as random cropping, horizontal flipping and color distortion. The initial weights of the ResNet18 used in the self-supervised task come from ImageNet pretraining as it is reported that this decision gives a performance boost in the downstream task \cite{Pantazis_2021_ICCV} while requiring less epochs to converge. Indicatively, for the experiments of this paper we train for 200 epochs and batch size 256.

To evaluate our proposed methodology, we use the linear evaluation protocol \cite{chen2020simple}, i.e. training a linear classifier for object classification, on top of the frozen layers of the feature extractor $f$ informed by self-supervised pretraining. The aforementioned procedure is typical for evaluating the quality of representations learnt by unsupervised approaches. The classifier is trained with 1\%, 10\% and 100\% of the available data to illustrate the point that self-supervised learning can do well in the lower data regime. These percentages correspond to number of training images that range between approximately 3,549 and 261,695 images, and can be important to illustrate the usefulness of the suggested approach across a range of data sizes. It's important to note that sampling takes place per class to maintain the class-imbalance as this exists in the original dataset.

\subsection{Results}

\subsubsection{Mining Positives with Polygon Matching}

As our main baseline, we used the three aforementioned standard self-supervised approaches with a ResNet18 \cite{he2016deep} as backbone. For completeness, we also train a ResNet18 as our supervised baseline initialized either randomly or with ImageNet pretrained weights.
Initially, we find out that self-supervised learning is a promising approach for learning high quality representation in images collected from robots navigating home environments. Specifically in Table \ref{tab:top1_acc}, we observe that the Top-1 accuracy boost we get by using the ``Standard'' self-supervised learning frameworks instead of ImageNet pretrained features is between $12\%$  and $17\%$, with the biggest gains recorded in the low data regime.  
In addition, our ``Polygon Matching'' suggestion for efficient self-supervised learning, further boosts the performance of the SSL paradigm, reporting gains consistently across all approaches and amounts of supervision. On average, the best results are reported with SimSiam, showing that negatives may not be necessary in a scenario where we the visual diversity of the selected positives has increased in a natural way such as matching multiple views of an object.

\begin{table}[h]
\centering
\footnotesize
\begin{tabular}{ |c|c|c|c|c| }
\hline
\multicolumn{5}{|c|}{\bf Top-1 Accuracy}\\
\hline
Approach & Method  & 1\% & 10\% & 100\% \\ 
\hline
Supervised
    & Random Init.  &37.7 &38.5 &39.4\\
 \hline
 
Supervised
    & ImageNet Init.  &57.8 &63.1 &65.2\\
\hline
SimCLR 
    & Standard &74.6 &75.7 &77.0 \\  
    & Polygon Matching &\textbf{75.0} &\textbf{76.5} &\textbf{77.5} \\  
 \hline
SimSiam
    & Standard  &74.0 &76.1 &77.8 \\ 
     & Polygon Matching &\textbf{75.5} &\textbf{77.7} &\textbf{78.5} \\  
\hline
Triplet
    & Standard  &62.8 &68.3 &70.3 \\ 
     & Polygon Matching &\textbf{70.6} &\textbf{73.1} &\textbf{73.9} \\  
\hline
\end{tabular}
\caption{Top-1 accuracy of ``Standard'' and ``Polygon Matching'' variants of the SSL along with supervised baselines of the same architecture (ResNet18) either initialized randomly or with ImageNet features. The reported performance corresponds to the accuracy after linearly evaluating with various amounts of supervision. Gains are reported for both vanila SSL when compared with an established transfer-learning baseline and the ``Polygon Matching'' positive mining technique when compared to ``Standard'' approaches.}
\label{tab:top1_acc}
\end{table}

\subsubsection{Consistency of Gains across Classes}

Moreover, the improvements achieved with ``Polygon Matching'' versus ``Standard'' are more significant when averaged per class. 
Given the imbalances that exist in any real-world dataset such as \emph{Vacuum Objects}, we want to make sure that the reported gains do not only represent a boost in the majority classes. From the results in Table \ref{tab:balanced_top1_acc}, we see that the increase in the downstream task performance is again consistent across all techniques and amounts of data. An assumption for the aforementioned finding is that by taking into account different perspectives of the same object during the self-supervised task helps especially the underrepresented classes, i.e. when a class is rarely observed we can make sure that we exploit all of the captured perspectives in the best way possible.

\begin{table}[h]
\centering
\footnotesize
\begin{tabular}{ |c|c|c|c|c| }
\hline
\multicolumn{5}{|c|}{\bf Balanced Top-1 Accuracy}\\
\hline
Approach & Method  & 1\% & 10\% & 100\% \\ 
\hline
SimCLR 
    & Standard &52.0 &53.5 &55.0 \\  
    & Polygon Matching &\textbf{53.1} &\textbf{54.5} &\textbf{56.4} \\  
 \hline
SimSiam
    & Standard  &50.5 &54.5 &56.5 \\ 
     & Polygon Matching &\textbf{53.0} &\textbf{56.7} &\textbf{58.2} \\  
\hline
Triplet
    & Standard  &41.9 &46.6 &48.7 \\ 
     & Polygon Matching &\textbf{48.7} &\textbf{50.8} &\textbf{52.8} \\  
\hline
\end{tabular}
\caption{Balanced Top-1 accuracy of ``Standard'' and ``Polygon Matching'' variants of the SSL approaches after linearly evaluating with various amounts of supervision. The performance improvements are not only consistent across the underlying techniques but also of greater magnitude compared to the overall Top-1 Accuracy discussed above.}
\label{tab:balanced_top1_acc}
\end{table}

\subsubsection{The Impact of the Maximum Depth considered}
\begin{figure}[t]
\begin{center}
 \includegraphics[width=0.6\columnwidth]{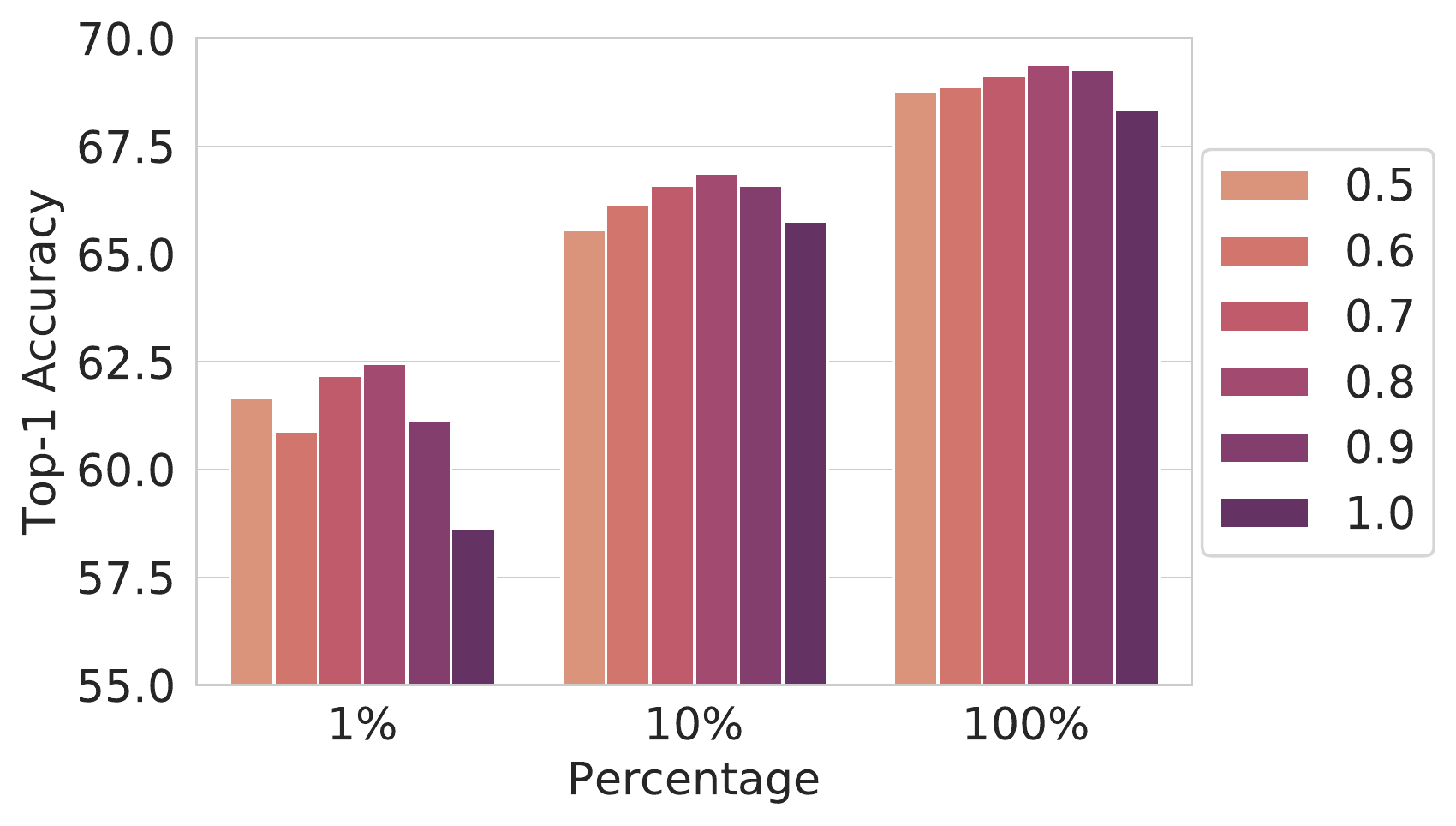}
\end{center}
   \caption{We varied the maximum distance away from of the camera that we consider for our suggested ``Polygon Matching'' approach. Top-1 Accuracy of the maximum distance variants ranging from 0.5 to 1.0 show that setting the maximum depth parameter around 0.7-0.8 meters gives the best performance.
   }
\label{fig:max_depth}
\end{figure}

When a mobile agent goes around and captures images, there can be cases where the objects detected lie far away from the camera. The suggested `Polygon Matching'' approach for mining positives can be parameterized with the maximum distance (meters) away from the camera that is considered during candidate selection. 
It is fair to assume that considering images captured in distance, will give us more candidate positives for self-supervised learning. 
We examine the effect of the maximum depth by varying its value between 0.5 and 1 meter away from the center of the robot. As we can see in Figure \ref{fig:max_depth}, the best performance across various levels of supervision is observed when the maximum distance is about 0.7-0.8 meters away from the camera. Thus, we can infer that considering objects that are too close to the camera may not give us enough positives, while a large threshold can hurt performance, maybe because of a the potential increase in the amount of false positives. For computational purposes, we sampled a smaller amount of training images for the experiments investigating the effect of maximum depth compared to other experiments in the paper and that is why the results in Figure \ref{fig:max_depth} should be observed independently.

\section{Conclusion}

In this work, we examined and proved the potential of self-supervised learning for efficient representation learning for data collected under the setting of a navigating robotic agent with an embedded camera. In addition, we demonstrated the effectiveness of retrieving more informative positives for self-supervised learning under the aforementioned setting by simply exploiting camera and robot information that typically comes with any navigating mobile robot. In particular, we show that a self-supervised learning task that learns to push together different perspectives of the same object instead of simply relying on self-augmentations leads to consistent gains across various representation learning techniques. Given the fact that the boost becomes even more significant when we average across classes, we expect our findings to be useful in real-world imbalanced data scenarios where an agent collects vast amounts of data and the annotation budget is limited. Finally, we believe that similar approaches can aid holistic understanding of objects, especially on tasks when on top of the object's variations we have to deal with a huge variety of real-world settings.

\clearpage
\bibliographystyle{splncs04}
\bibliography{main}
\end{document}